%% file: main.tex
\documentclass[conference,letterpaper,10pt]{ieeeconf}

\usepackage{cite}
\usepackage{amsmath,amssymb,amsfonts}
\usepackage{graphicx}
\usepackage{textcomp}
\usepackage{xcolor}
\usepackage{booktabs}
\usepackage{url}
\usepackage[hidelinks]{hyperref}

\author{\IEEEauthorblockN{Rafiqul Islam}
\IEEEauthorblockA{Independent Researcher\\
Email: rafiqul713@gmail.com}}

\title{When Faster VLA Deployment Changes Closed-Loop Behavior: Task Success--Latency Analysis of SmolVLA Across PyTorch and ONNX Variants}

\begin{document}
\maketitle

\begin{abstract}
Vision-language-action (VLA) deployment can reduce inference latency while changing closed-loop task behavior.
We evaluate \texttt{HuggingFaceVLA/smolvla\_libero} on an RTX~2060 (6\,GB) in LIBERO Spatial and Object (MuJoCo 3.3.2, LeRobot 0.6.1, seed~42), comparing PyTorch+AMP with ONNX Runtime CUDA Execution Provider (CUDA EP).
The main evaluation uses 100 episodes/suite; a paired rollout uses 300 episodes/suite.
PyTorch+AMP reaches 70.0\%/88.0\% Spatial/Object success at 1181\,ms p99.
ONNX export invocations requested as FP16 and INT8 yield tether-inspect p99 values of 601\,ms and 532\,ms, while Spatial success falls to 41.0\% and 40.0\% and Object remains at 89.0\%; graph audit shows both artifacts are byte-identical FP32 graphs, so the requested-INT8 result is not operator-level INT8 quantization.
A static language-width ablation (16/24/32 tokens) yields Spatial success of 41.0\%, 75.0\%, and 71.0\%; widths 24 and 32 recover much of the Spatial drop while Object success and uniform-bench latency stay approximately stable.
Width-24 ONNX Spatial success was 75.0\%, close to the PyTorch+AMP baseline of 70.0\%; their Wilson 95\% confidence intervals overlap.
Context width is an important contributor in this stack; it does not account for every PyTorch-vs-ONNX difference.
Deployment evaluation should jointly report latency, artifact inspection, interface constraints, and closed-loop success.
Code: \url{https://github.com/rafiqul713/smolvla-libero-onnx}.
\end{abstract}


\input{sections/01_introduction}
\input{sections/02_method}
\input{sections/03_results}
\input{sections/04_discussion}
\input{sections/05_conclusion}

\section*{Acknowledgment}
The author thanks the Hugging Face LeRobot and SmolVLA teams for open checkpoints and simulation benchmarks.
All experiments used personal hardware (NVIDIA RTX~2060); no employer resources were used.

\bibliographystyle{IEEEtran}
\bibliography{references}

\end{document}

%% file: sections/01_introduction.tex
\section{Introduction}
\label{sec:introduction}

SmolVLA~\cite{shukor2025smolvla} reports strong LIBERO scores under floating-point PyTorch evaluation.
A typical deployment path then changes several variables together: ONNX export, runtime, requested precision, kernels, and static input shapes.
The question in this letter is not only whether inference becomes faster, but whether \emph{closed-loop task success} is preserved.

Three quantities that are often treated as interchangeable are distinct:
(i)~open-loop numerical agreement on shared inputs;
(ii)~latency and memory on a given GPU;
(iii)~closed-loop success, where each action changes the next state and small output differences can accumulate.
High values of (i) or (ii) do not imply (iii).

OpenVLA~\cite{kim2024openvla} and $\pi_0$~\cite{black2024pi0} established open VLA policies; NanoVLA~\cite{nanovla2025} and LiteVLA-Edge~\cite{williams2026litevlaedge} target smaller or on-device policies.
QuantVLA~\cite{zhang2026quantvla} and EaqVLA~\cite{eaqvla2025} study post-training quantization, typically on datacenter GPUs.
vla.cpp~\cite{vlacpp2026} reports SmolVLA latency on Jetson via GGUF without coupling those timings to LIBERO success under ONNX-style exports.
This letter does not propose a new policy or quantizer.
We study a narrower setting: \texttt{HuggingFaceVLA/smolvla\_libero} on one RTX~2060 (6\,GB), with a pinned LIBERO Spatial/Object protocol (MuJoCo 3.3.2, LeRobot 0.6.1, seed~42).
Exports use tether~\cite{tether2026}; ONNX latency uses CUDA Execution Provider (CUDA EP).
TensorRT engine compilation was attempted but did not complete within 6\,GB; we do not report TensorRT latency.

The main observation is a deployment gap: the ONNX/CUDA EP configuration is substantially faster than PyTorch+AMP on this GPU (about $2\times$ lower p99 on the respective benches) but shows substantially lower Spatial closed-loop success, while Object success remains comparable.
The export tool's default 16-token language width is a testable interface hypothesis for part of this asymmetry.
We evaluate it with a 16/24/32-token re-export, holding checkpoint, tether version, runtime, hardware, and evaluation protocol fixed.

\paragraph{Contributions}
\begin{enumerate}
    \item Empirical evidence that a faster ONNX+CUDA-EP deployment configuration of SmolVLA need not preserve closed-loop LIBERO Spatial success, while Object success remains comparable (100-episode/suite main evaluation, seed~42).
    \item A controlled 16/24/32-token language-width ablation, with checkpoint, tether version, CUDA EP, hardware, and protocol held fixed, providing evidence that static language-context width is an important contributor to the Spatial gap in the tested stack.
    \item A graph-level audit showing that a requested precision flag need not imply actual conversion: the requested-INT8 artifact matches the requested-FP16 FP32 graph and contains no quantization operators.
    \item Evidence that open-loop numerical agreement does not guarantee closed-loop task-success agreement, including a 300-episode/suite paired PyTorch--ONNX rollout (Spatial 56.7\% vs.\ 33.0\%; Object 73.3\% vs.\ 74.0\%) and an early adapter with $\sim$0.98 per-step cosine similarity but 0\% closed-loop success.
\end{enumerate}

Our PyTorch+AMP baseline (70\% Spatial, 88\% Object) is below the SmolVLA paper (96\%/92\%)~\cite{shukor2025smolvla}.
This letter does not attempt to reproduce those headline rates; relative comparisons use one local stack.
Code and measured results: \url{https://github.com/rafiqul713/smolvla-libero-onnx}.

%% file: sections/02_method.tex
\section{Experimental Setup}
\label{sec:method}

\subsection{Hardware, software, and protocol}

All runs used a Linux workstation with an NVIDIA RTX~2060 (6\,GB, Turing sm\_75), driver 535.309.01, Python 3.12, LeRobot 0.6.1, PyTorch 2.5.1+cu121, MuJoCo 3.3.2 (\texttt{MUJOCO\_GL=egl}), tether 0.12.0, and ONNX Runtime GPU 1.22.0.
No Jetson device and no physical robot were used.

SmolVLA~\cite{shukor2025smolvla} combines a SigLIP vision tower, a SmolLM2 backbone, and a ten-step flow-matching action expert (7-DoF actions).
LIBERO~\cite{liu2023libero} has four suites of ten tasks; we evaluate Spatial and Object~\cite{shukor2025smolvla,lerobot2026libero}.
Scores are sensitive to MuJoCo version and action-step settings~\cite{lerobotissue3264}.
Table~\ref{tab:eval_protocol} lists the pinned protocol.
PyTorch evaluation uses \texttt{load\_vlm\_weights=false} (weights from the LIBERO checkpoint; load-time only) and \texttt{use\_amp=true} (\texttt{torch.autocast} defaults to float16 on CUDA in PyTorch 2.5).
We therefore call this the \emph{PyTorch+AMP baseline}, not pure FP32 inference.

\begin{table}[t]
    \centering
    \caption{Pinned protocol. Main evaluation and paired rollouts differ in episode count and harness.}
    \label{tab:eval_protocol}
    \resizebox{\columnwidth}{!}{%
    \begin{tabular}{ll}
        \toprule
        Setting & Value \\
        \midrule
        Checkpoint & \texttt{HuggingFaceVLA/smolvla\_libero} \\
        MuJoCo / GL & 3.3.2 / \texttt{egl} \\
        Action / flow steps & 1 / 10 \\
        Tasks per suite / batch / seed & 10 / 1 / 42 \\
        Main evaluation & 10 eps/task (100 eps/suite) \\
        Paired rollout evaluation & 30 eps/task (300 eps/suite) \\
        PyTorch AMP & on \\
        GPU / ONNX runtime & RTX 2060 6\,GB / CUDA EP \\
        Requested ONNX precision & fp16, int8 (Sec.~\ref{sec:audit}) \\
        Language widths & 16 / 24 / 32 tokens \\
        \bottomrule
    \end{tabular}}
\end{table}

\subsection{Deployment variants and measurements}

Variants: (i)~PyTorch+AMP; (ii)~ONNX from \texttt{tether export --precision fp16}; (iii)~ONNX from \texttt{--precision int8}; (iv)~the same checkpoint re-exported at language widths 24 and 32.
Default tether traces language inputs at width 16.
Requested precision is a flag; actual dtypes are established by graph audit (Section~\ref{sec:audit}), not by the flag.

Latency uses 50 warmup + 1000 timed calls, batch 1, and ten flow steps, with CUDA synchronization around each call.
Two ONNX benches exist and are not interchangeable: \texttt{tether inspect bench} (Table~\ref{tab:summary}) and a uniform ONNX Runtime script used for all three widths (Table~\ref{tab:ctx}, Fig.~\ref{fig:pareto}).
PyTorch latency uses a separate AMP bench.
Reported latency values characterize those respective configurations and provide an empirical deployment comparison, not a fully controlled cross-runtime microbenchmark.
Peak GPU memory is process-level \texttt{nvidia-smi} during inference (indicative process-level footprint, not allocator peak).
Open-loop agreement: cosine similarity and maximum absolute difference of action chunks (first 7 dimensions) on a stated input set.
Closed-loop success: LIBERO episode success (\%).
We treat these as distinct metrics.

Main closed-loop runs use \texttt{lerobot-eval} / \texttt{scripts/libero\_onnx\_eval.py} (100 episodes/suite).
Paired PyTorch--ONNX runs use \texttt{scripts/run\_smolvla\_verify.py} on the same tether control loop (300 episodes/suite) to compare behavior under matched seeds and resets; they do not isolate a single deployment factor.
An early adapter re-tokenized instructions with the export tokenizer and scored 0\% at $\sim$0.98 per-step cosine vs.\ PyTorch.
The corrected adapter passes LeRobot \texttt{lang\_tokens} and camera masks, padded or truncated to the graph width.

TensorRT EP compilation was attempted after stripping unsupported ScatterND attributes and did not finish within 6\,GB.
All reported ONNX latency uses CUDA EP.

\subsection{Context-length ablation}
\label{sec:ablation-protocol}

LeRobot appends a newline and tokenizes with the SmolVLM2 tokenizer (\texttt{max\_length=48}).
Some Spatial instructions then exceed 16 tokens at the ONNX boundary.
Widths 24 and 32 cover the longest Spatial instruction in this suite (20 tokens including newline).
Each width is a separate tether 0.12.0 export (same patches, opset 19, ten denoise steps).
Fixed: hardware, CUDA EP, seed 42, 100-episode main protocol, and LeRobot tokens.

Per width we measure: truncation frequency (tasks whose \emph{fed} token count exceeds the static width), Spatial/Object success, and uniform-bench p50/p99 and peak memory.
A supporting outcome is Spatial recovery at width $\ge$24 with Object and latency approximately stable.
A weakening outcome is no Spatial recovery, in which case Table~\ref{tab:factors} lists the remaining candidates.
Either outcome is informative.

%% file: sections/03_results.tex
\section{Results}
\label{sec:results}

\subsection{Task success and latency under the main evaluation}

Table~\ref{tab:summary} reports the 100-episode/suite main evaluation (seed~42).
PyTorch+AMP: 70.0\% Spatial, 88.0\% Object, p99 1181\,ms, p50 903\,ms, mean 917\,ms, peak process memory 2208\,MB (\texttt{nvidia-smi}).
Object is within a few points of the SmolVLA paper (92\%); Spatial is substantially lower (96\% in~\cite{shukor2025smolvla}).
We do not attribute that discrepancy to a single factor.

Requested-FP16 and requested-INT8 ONNX (width 16) yield \texttt{tether inspect bench} p99 601\,ms and 532\,ms, Spatial 41.0\% and 40.0\%, and Object 89.0\% in both runs.
The uniform ONNX Runtime bench used later for width comparison (Table~\ref{tab:ctx}) reports width-16 p99 583.6\,ms; 601\,ms versus 583.6\,ms reflects harness differences (\texttt{tether inspect bench} versus direct ORT \texttt{Session.run}), not a different model.
Relative to PyTorch+AMP, p99 is about $2\times$ lower, Spatial drops by 29--30 percentage points, and Object stays within one point.
These timings are CUDA EP, not TensorRT.
ONNX peak process memory is about 3.2\,GB (requested-INT8 VRAM was not recorded).
Disabling AMP raises PyTorch p99 to 1261\,ms (supporting measurement; not the baseline).

The PyTorch$\rightarrow$ONNX step changes framework, graph, arithmetic (AMP kernels vs.\ an FP32 graph), kernels, and runtime together.
The speedup characterizes that coupled configuration, not a single precision switch.

\begin{table}[t]
    \centering
    \caption{Main 100-episode/suite evaluation (seed 42, width 16). Success is LIBERO episode success (\%); brackets are Wilson 95\% CIs ($n{=}100$). ONNX p99: \texttt{tether inspect bench} (ms). VRAM: process-level \texttt{nvidia-smi} (MB), not allocator peak.}
    \label{tab:summary}
    \resizebox{\columnwidth}{!}{%
    \begin{tabular}{llcccc}
        \toprule
        Runtime & Requested & Spatial & Object & p99 & VRAM \\
        \midrule
        PyTorch+AMP & AMP & 70.0\,[60.4, 78.1] & 88.0\,[80.2, 93.0] & 1181 & 2208 \\
        ONNX CUDA EP & fp16 & 41.0\,[31.9, 50.8] & 89.0\,[81.4, 93.7] & 601 & 3200 \\
        ONNX CUDA EP & int8 & 40.0\,[30.9, 49.8] & 89.0\,[81.4, 93.7] & 532 & --- \\
        SmolVLA paper~\cite{shukor2025smolvla} & FP32 & 96.0 & 92.0 & --- & --- \\
        \bottomrule
    \end{tabular}}
    {\footnotesize Requested-fp16/int8 rows are export invocations, not verified precision conversions (Section~\ref{sec:audit}). No TensorRT row. Latency is not a fully controlled PyTorch-vs-ONNX microbenchmark.}
\end{table}

\subsection{Open-loop identity check and 300-episode paired rollouts}
\label{sec:parity}

On 200 shared random inputs (seed 42), the requested-FP16 and requested-INT8 ONNX artifacts produce cosine similarity 1.0000 and zero maximum absolute difference at the recorded precision.
This is an identity/parity check between those two artifacts on that input set.
It is not a PyTorch-vs-ONNX comparison, and it is not evidence that INT8 quantization preserves FP16 outputs.

Paired rollouts compare native PyTorch with the width-16 ONNX export on the same tether loop (30 episodes/task $\times$ 10 tasks $=$ 300 episodes/suite, seed 42; Table~\ref{tab:verify}).
PyTorch Spatial/Object: 56.7\%/73.3\%; ONNX: 33.0\%/74.0\%.
Object is near-matched; Spatial shows a large ONNX deficit.
Absolute rates differ from the 100-episode main evaluation because the harnesses differ; the Spatial-vs-Object pattern is the same.
The pairing reduces some seed/reset variation but does not isolate one runtime factor.
An earlier adapter with $\sim$0.98 per-step cosine vs.\ PyTorch still scored 0\% closed-loop success, which is consistent with open-loop numerical agreement not implying closed-loop task agreement.

\begin{table}[t]
    \centering
    \caption{Paired 300-episode/suite evaluation (30 episodes/task, seed 42, same tether loop). Not the 100-episode main evaluation. McNemar: continuity-corrected $\chi^2$ on paired per-episode success/failure.}
    \label{tab:verify}
    \resizebox{\columnwidth}{!}{%
    \begin{tabular}{lccccc}
        \toprule
        Suite & PyTorch & ONNX (width 16) & Gap (pp) & McNemar $\chi^2$ & $p$ \\
        \midrule
        Spatial & 56.7 & 33.0 & $-23.7$ & 35.77 & $2.2\times10^{-9}$ \\
        Object & 73.3 & 74.0 & $+0.7$ & 0.014 & 0.90 \\
        \bottomrule
    \end{tabular}}
\end{table}

\subsection{Graph audit of requested FP16 and INT8 artifacts}
\label{sec:audit}

The requested-FP16 and requested-INT8 artifacts were byte-identical and contained FLOAT initializers without quantization operators (same SHA-256 for \texttt{model.onnx} and the 2.3\,GB weight file; 22\,514 nodes).
Consequently, the requested-INT8 result does not represent operator-level INT8 quantization, and the observed latency difference between the two runs (601 vs.\ 532\,ms p99) cannot be attributed to INT8 quantization.
That difference is consistent with run-to-run CUDA-EP/runtime variability; the present experiments do not isolate its cause.
tether 0.12.0's monolithic path does not apply \texttt{--precision}; the flag is consumed by the TensorRT build path, which did not complete within 6\,GB, so no TensorRT INT8 latency is reported.
The 200-input identity check is consistent with the audit and does not establish robustness of a genuine INT8 implementation.
Spatial 41.0\% vs.\ 40.0\% (one episode in 100) bounds success noise for this configuration.

\subsection{Static language-context ablation}
\label{sec:ablation}

\begin{figure}[t]
    \centering
    \includegraphics[width=\columnwidth]{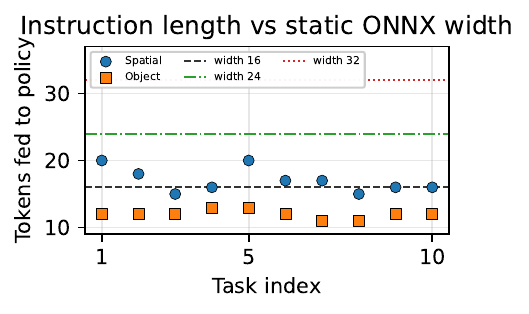}
    \caption{Fed instruction lengths (newline included) for ten Spatial and ten Object tasks versus static ONNX widths. Spatial: 5/10 exceed 16, max 20. Object: 0/10 exceed 16, max 13.}
    \label{fig:tokens}
\end{figure}

LeRobot appends a newline and tokenizes with the SmolVLM2 tokenizer (\texttt{max\_length=48}).
\emph{Fed} counts include that newline; \emph{raw} counts omit it and are one token shorter.
Spatial raw lengths are approximately 15--19 tokens (mean 16.0, 3/10 exceed 16, max 19); fed lengths are 15--20 tokens (mean 17.0), and 5 of 10 Spatial tasks exceed width 16 (tasks 0, 1, 4, 5, 6).
Object fed lengths are 11--13 tokens (mean 12.0); none exceed 16.
No instruction in these two suites exceeds 24 tokens (Fig.~\ref{fig:tokens}).
Tables and grouping use fed counts.

Table~\ref{tab:pertask} groups Spatial episodes by whether the task was truncated \emph{at width 16} (five tasks, 50 episodes).
That grouping is fixed across rows; it does not mean the same tasks are truncated at widths 24 or 32.
With PyTorch the two groups are comparable (38/50 vs.\ 32/50).
With width-16 ONNX, previously-truncated-at-16 tasks fall to 3/50 and 4/50 (requested-fp16/int8 runs) while untruncated-at-16 tasks remain 38/50 and 36/50.

Table~\ref{tab:ctx} reports the 16/24/32 experiment (100 episodes/suite; uniform CUDA-EP bench).
Width 24 raises Spatial success from 41.0\% (requested-FP16 width 16) to 75.0\%; width 32 yields 71.0\%.
Width-24 ONNX Spatial success was 75.0\%, close to the PyTorch+AMP baseline of 70.0\%; their Wilson 95\% confidence intervals overlap.
A two-proportion $\chi^2$ comparing these rates ($p{=}0.53$) does not establish equivalence.
Width 24 therefore substantially recovers Spatial performance; width 32 does not show an additional improvement over width 24.
The 75.0\% vs.\ 71.0\% difference should not be read as a statistically established width effect.
Recovery is concentrated in previously-truncated-at-16 tasks (3--4/50 $\rightarrow$ 36/50 at width 24; 34/50 at width 32); untruncated-at-16 tasks stay in 36--39/50.
Object success remains approximately stable (89.0\%, 90.0\%, 87.0\%).
Uniform-bench p50/p99 and peak memory change only slightly (Table~\ref{tab:ctx}).

This provides evidence that static language-context width is an important contributor to the Spatial gap in the tested stack.
It does not show that width accounts for every PyTorch-vs-ONNX difference, and it does not establish a universal optimal width.

On matched random images, state, and noise, width-16 and width-24 graphs agree on an 11-token Object instruction (cosine 1.000000, max $|\Delta|$ $3.8\times10^{-6}$).
On a 19-token encoding of Spatial task 0, width 16 differs from PyTorch \texttt{sample\_actions} (cosine 0.952, max $|\Delta|$ 1.73) whereas width 24 agrees more closely (cosine 0.99997, max $|\Delta|$ 0.028).
This is consistent with truncation changing the model input; it is one-instruction evidence, not a suite-wide open-loop test.

\begin{figure}[t]
    \centering
    \includegraphics[width=\columnwidth]{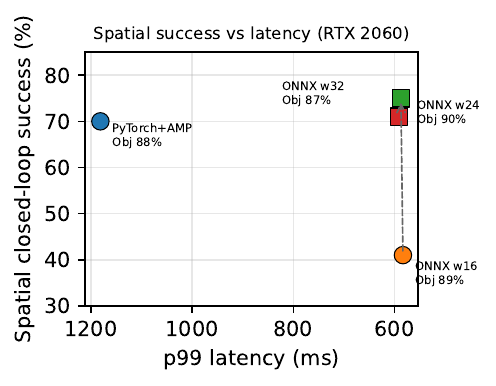}
    \caption{p99 latency versus Spatial closed-loop success (Object annotated). ONNX widths use the uniform CUDA-EP bench of Table~\ref{tab:ctx}; PyTorch uses the separate AMP bench. These harnesses are not identical. The 601/532\,ms tether-inspect values are not plotted.}
    \label{fig:pareto}
\end{figure}

Fig.~\ref{fig:pareto} uses the \emph{same} uniform ONNX bench as Table~\ref{tab:ctx} for all three widths (PyTorch p99 remains the separate AMP bench).
Width-24/32 occupy the same latency region as width 16 but recover Spatial success, which is consistent with the observed Spatial degradation being partly attributable to the static language interface rather than to the small latency differences across the tested ONNX widths.

\begin{table}[t]
    \centering
    \caption{Spatial episodes solved of 50. Columns use a fixed grouping: tasks truncated at width 16 vs.\ tasks not truncated at width 16. Widths 24 and 32 truncate none of these tasks. Brackets: Wilson 95\% CI on the $n{=}50$ proportion (\%).}
    \label{tab:pertask}
    \resizebox{\columnwidth}{!}{%
    \begin{tabular}{lccc}
        \toprule
        Configuration & Prev.\ trunc.\ at 16 & Untrunc.\ at 16 & Total /100 \\
        \midrule
        PyTorch+AMP & 38\,[62.6, 85.7] & 32\,[50.1, 75.9] & 70 \\
        ONNX w16 (req.\ fp16) & 3\,[2.1, 16.2] & 38\,[62.6, 85.7] & 41 \\
        ONNX w16 (req.\ int8) & 4\,[3.2, 18.8] & 36\,[58.3, 82.5] & 40 \\
        ONNX w24 & 36\,[58.3, 82.5] & 39\,[64.8, 87.2] & 75 \\
        ONNX w32 & 34\,[54.2, 79.2] & 37\,[60.4, 84.1] & 71 \\
        \bottomrule
    \end{tabular}}
\end{table}

\begin{table*}[t]
    \centering
    \caption{Context-length ablation (100 episodes/suite, seed 42, uniform CUDA-EP bench: 50 warmup, 1000 calls). Truncation uses fed token counts. Width-16 success is the requested-fp16 export; the requested-int8 replicate is 40.0\%/89.0\%. Success brackets: Wilson 95\% CIs ($n{=}100$).}
    \label{tab:ctx}
    \begin{tabular}{cccccccc}
        \toprule
        Width & Spatial trunc. & Object trunc. & Spatial suc. & Object suc. & p50 (ms) & p99 (ms) & Peak mem.\ (MB) \\
        \midrule
        16 & 5/10 & 0/10 & 41.0\,[31.9, 50.8] & 89.0\,[81.4, 93.7] & 537.5 & 583.6 & 3184 \\
        24 & 0/10 & 0/10 & 75.0\,[65.7, 82.5] & 90.0\,[82.6, 94.5] & 544.5 & 587.7 & 3178 \\
        32 & 0/10 & 0/10 & 71.0\,[61.5, 79.0] & 87.0\,[79.0, 92.2] & 549.2 & 591.3 & 3178 \\
        \bottomrule
    \end{tabular}
\end{table*}

%% file: sections/04_discussion.tex
\section{Discussion}
\label{sec:discussion}

\subsection{What the measurements support}

Faster VLA inference is not automatically behavior-preserving.
The ONNX/CUDA EP configuration is substantially faster than the PyTorch+AMP bench on this GPU (roughly half the measured p99 latency on the respective benches), yet default-width Spatial closed-loop success falls by about 30 percentage points in the 100-episode main evaluation while Object does not.
Those latency figures characterize the respective benches, not a fully controlled microbenchmark.
The 200-input ONNX-vs-ONNX identity check does not detect the closed-loop Spatial change; it only confirms that the two requested-precision artifacts match on those inputs.

Increasing width from 16 to 24 substantially improves Spatial success (41.0\%/40.0\% $\rightarrow$ 75.0\%); width 32 (71.0\%) does not show an additional benefit.
Object success (89.0\% $\rightarrow$ 90.0\%/87.0\%) and uniform-bench p99 (583.6 $\rightarrow$ 587.7/591.3\,ms) remain approximately stable.
Truncation falls from 5/10 Spatial tasks to 0/10, and recovery is concentrated in previously-truncated-at-16 tasks (Table~\ref{tab:pertask}).
This provides evidence that static language-context width is an important contributor to the Spatial gap in the tested stack.
It does not show that width accounts for every PyTorch-vs-ONNX difference, or that the same failure appears for other exporters or VLAs.

\subsection{Coupled deployment factors}

Table~\ref{tab:factors} records isolation status.
Changing from PyTorch+AMP to ONNX/CUDA EP can jointly change runtime, graph representation, kernels, numerical execution, language-context handling, memory behavior, and scheduling.
The width ablation is useful because it varies one interface parameter while holding checkpoint, tether version, CUDA EP, hardware, and protocol fixed.
The letter does not perform a full factorial isolation of PyTorch vs.\ ONNX, AMP vs.\ non-AMP, graph conversion, CUDA-EP kernels, language width, true FP16/INT8 conversion, or TensorRT.
The requested-INT8 export is a second invocation of an unconverted graph (Section~\ref{sec:audit}).
TensorRT execution was not obtained.
We therefore do not claim complete causal attribution for the PyTorch-vs-ONNX gap, and we do not claim that latency itself caused the Spatial degradation.

\begin{table}[t]
    \centering
    \caption{Deployment factors evaluated in controlled ablations.}
    \label{tab:factors}
    \begin{tabular}{ll}
        \toprule
        Factor & Isolated? \\
        \midrule
        Language width (16/24/32) & Yes \\
        PyTorch vs.\ ONNX runtime & No \\
        Graph conversion & No \\
        AMP vs.\ FP32 graph & No \\
        CUDA-EP kernels & No \\
        True INT8 vs.\ FP16 conversion & No \\
        TensorRT engine execution & No \\
        \bottomrule
    \end{tabular}
\end{table}

A plausible interpretation, consistent with the observed pattern, is that truncating Spatial instructions can remove task-disambiguating information near the instruction tail, whereas Object instructions are shorter.
This reading was not independently tested.

\subsection{Limitations}

\begin{itemize}
    \item One GPU (RTX~2060 6\,GB); simulation only; no physical robot; no edge-device latency.
    \item LIBERO Spatial and Object only; Goal and Long are unevaluated.
    \item One checkpoint; one main seed (42); main evaluation 10 episodes/task; paired evaluation 30 episodes/task. Suite-level Wilson 95\% CIs and paired McNemar tests are reported; we do not run a multi-seed analysis.
    \item Reported significance tests (paired McNemar; independent two-proportion $\chi^2$ for width-24 vs.\ PyTorch+AMP Spatial) are presented individually without multiple-comparison correction. A non-significant two-proportion test is not an equivalence result.
    \item TensorRT compilation/benchmarking was not completed under the 6\,GB constraint.
    \item The requested-INT8 artifact is not operator-level INT8; the 200-input check compares those identical ONNX files, not PyTorch vs.\ ONNX.
    \item Factors other than language width are not factorially isolated (Table~\ref{tab:factors}).
    \item Width ablation tests only 16/24/32 tokens in one exporter/toolchain.
    \item The PyTorch+AMP Spatial score (70\%) differs from the SmolVLA publication (96\%) and is not a direct reproduction of that headline rate.
    \item Results do not generalize to all VLAs, exporters, or platforms.
\end{itemize}

%% file: sections/05_conclusion.tex
\section{Conclusion}
\label{sec:conclusion}

ONNX/CUDA EP produced substantially lower measured p99 latency than PyTorch+AMP in the tested deployment configurations (1181\,ms $\rightarrow$ 601/532\,ms on the respective benches).
That faster configuration did not preserve Spatial closed-loop success under the default width-16 export (70.0\% $\rightarrow$ 41.0\%/40.0\% in the 100-episode main evaluation); Object success remained approximately stable (88.0\% vs.\ 89.0\%).
The 300-episode paired evaluation showed the same qualitative Spatial/Object pattern (56.7\% vs.\ 33.0\%; 73.3\% vs.\ 74.0\%).
Increasing static language-context width to 24 recovered Spatial success (75.0\%); width 32 (71.0\%) did not show an additional benefit, while Object success and uniform-bench latency stayed approximately stable.
Width-24 ONNX Spatial success was 75.0\%, close to the PyTorch+AMP baseline of 70.0\%; their Wilson 95\% confidence intervals overlap.
The requested-INT8 artifact was not operator-level INT8 quantized---it matched the requested-FP16 FP32 graph---so artifact-level verification is important before attributing observed behavior or latency differences to numerical precision.

Deployment evaluation should jointly consider system-level latency and closed-loop task success, together with interface constraints and exported-graph inspection.
These measurements are for one model, one GPU, and two LIBERO suites.
Artifacts: \url{https://github.com/rafiqul713/smolvla-libero-onnx}.